\documentclass[nonacm,acmlarge]{acmart}

\makeatletter
\def\@ACM@checkaffil{
    \if@ACM@instpresent\else
    \ClassWarningNoLine{\@classname}{No institution present for an affiliation}%
    \fi
    \if@ACM@citypresent\else
    \ClassWarningNoLine{\@classname}{No city present for an affiliation}%
    \fi
    \if@ACM@countrypresent\else
        \ClassWarningNoLine{\@classname}{No country present for an affiliation}%
    \fi
}
\makeatother

\title{RepDL: Bit-level Reproducible Deep Learning Training and Inference}
\author{Peichen Xie}
\author{Xian Zhang}
\author{Shuo Chen}
\affiliation{Microsoft Research}
\authorsaddresses{}

\begin{document}

\begin{abstract}
Non-determinism and non-reproducibility present significant challenges in deep learning, leading to inconsistent results across runs and platforms. These issues stem from two origins: random number generation and floating-point computation. While randomness can be controlled through deterministic configurations, floating-point inconsistencies remain largely unresolved. To address this, we introduce RepDL, an open-source library that ensures deterministic and bitwise-reproducible deep learning training and inference across diverse computing environments. RepDL achieves this by enforcing correct rounding and order invariance in floating-point computation. The source code is available at \url{https://github.com/microsoft/RepDL}.
\end{abstract}

\maketitle

\section{Introduction}

Non-determinism and non-reproducibility are recognized problems in deep learning training and inference \cite{pham_problems_2020,summers_nondeterminism_2021,liu_reproducibility_2022,zhuang_randomness_2022}. Non-determinism refers to the phenomenon where repeated executions of the same task yield different results given identical inputs and systems. This is also known as run-to-run inconsistency. Compared to non-determinism, non-reproducibility is a more challenging issue where executions of the same task given identical inputs on various systems yield different results. This is also known as cross-platform inconsistency. These issues complicate the deployment and debugging of models in production environments, impair the correctness of cross-platform applications, undermine confidence in published results, and affect the trustworthiness of AI systems in sensitive domains \cite{gundersen_state_2018,hutson_artificial_2018}.

Addressing these issues requires a nuanced understanding of where non-determinism and non-reproducibility originate in modern systems. From a confluence of factors that span hardware, software, and algorithmic design, we identify two origins: the random number generator and the floating-point computation.

Most deep learning models use random number generators for random initialization, data shuffling, data augmentation, data sampling, etc. Unless explicitly managed, this randomness leads to different training and inference trajectories. However, even if the random number generator is configured to be deterministic and reproducible, non-determinism and non-reproducibility still exist in deep learning due to floating-point computation issues.

Compared to the random number generator, the floating-point computation is more intricate. First, mathematical functions such as the exponential function can be implemented with different algorithms and precisions, resulting in inconsistent numerical outputs across different systems. Second, as floating-point arithmetic is non-associative, different computation orders can lead to subtle (sometimes significant) numerical differences. When different systems execute the same deep learning task, different hardware, system configurations, compilers, libraries, and software implementations can introduce discrepancies in computation order. Even if the task is executed with the same system, parallelism of atomic operations and dynamic scheduling can also make computation orders inconsistent from run to run.

Although both industry and academia have proposed many measures to solve these problems, the numerical inconsistency is still an open problem in deep learning \cite{chen_towards_2022,xu_checkpointing_2022}. Focusing on this problem, we introduce RepDL, an open-source library enabling deterministic and reproducible deep learning. Specifically, RepDL ensures bitwise-consistent results  for deep learning training and inference across multiple executions with the same or different CPU or GPU systems.

To tackle the floating-point computation problem, we adhere to two principles in RepDL development: (1) correct rounding and (2) order invariance. RepDL implements the basic operations (including arithmetic operations, square root operation, exponential function, logarithm function, trigonometric functions, etc.) with correct rounding, ensuring consistent precision and bit-level equivalence. For order invariance, RepDL implements reduction operations such as summation, matrix multiplication, and convolution with fixed reduction order while retaining parallelism, and implements deep learning functions with fixed combination of basic operations. 

Building on reproducible operations, we provide APIs compatible with PyTorch. Specifically, RepDL supports deep learning operations, differentiable functions, neural network modules, and optimizers defined in PyTorch, keeping their names and parameter definitions intact. For example, \texttt{repdl.nn.conv2d} is the reproducible version of PyTorch's two-dimensional convolution layer \texttt{torch.nn.conv2d}. RepDL is open-sourced at \url{https://github.com/microsoft/RepDL} and encourages extensions and community contributions.
\section{Origins of Non-determinism and Non-reproducibility}

There are many causes of non-determinism and non-reproducibility in deep learning, and we classify them into two origins: the random number generator and floating-point computation.

\subsection{Random number generator}

Random number generators (RNGs) are used in deep learning for various purposes, such as initializing weights, shuffling data samples, applying dropout regularization, generating noise for data augmentation, and so on. However, RNGs can introduce significant variability due to different seeds, inconsistent RNG algorithms or non-deterministic call sequences in multithreaded environments.

There are several solutions to RNG-related issues by controlling the random number generator \cite{chen_towards_2022,xu_checkpointing_2022}. For example, one solution adopted by PyTorch is using a reproducible RNG algorithm (e.g., MT19937) in a thread-safe manner: each thread has an independent RNG and the local seed is calculated from a deterministic function of the base seed and the thread index. By adopting this mechanism and setting a fixed base seed, the random numbers generated by the RNG are reproducible.

\subsection{Floating-point computation}

Deep learning uses floating-point numbers to represent data and computational results. However, due to the intrinsic properties of floating-point arithmetic, different implementations of the same deep learning operation can lead to numerical discrepancies. In other words, if the operation were computed using integer or real numbers, there would be no discrepancy. 

We investigate implementation differences causing the numerical discrepancy, and categorize them into two types: (1) the precision of basic operations and (2) the computation order.

\subsubsection{Precision of basic operations}

Basic operations, including arithmetic operations and basic mathematical functions, can be implemented with different precision across different systems, making the numerical results non-reproducible. Both software and hardware can introduce discrepancies in numerical precision.

Most basic mathematical functions are implemented numerically inconsistent between different libraries \cite{innocente_accuracy_2023}. For example, the $\log x$ function has different precision between GNU libc (glibc) and Intel Math Library.

In addition, basic mathematical functions computed by hardware instructions also have different precisions across different hardware. For example, the precision of the reciprocal instruction RCP which calculates $1/x$ can vary between different x86 CPUs.

\subsubsection{Computation order}

Floating-point arithmetic is non-associative. Therefore, different orders of operations or transformations of mathematical expressions can lead to numerical discrepancy. For example, using single-precision floating-point (float32) numbers, $(0.5 + 10^9) - 10^9=0$ while $0.5 + (10^9 - 10^9)=0.5$, and $\sqrt{2\cdot x}\neq \sqrt{2}\cdot\sqrt{x}$. The computation orders of deep learning operations may be non-deterministic and non-reproducible due to many factors.

Non-deterministic factors include:

\begin{itemize}
    \item \textbf{Atomic operations.} Reduction operations in deep learning can be implemented using atomic operations. For example, some convolution and scatter-add implementations are based on the atomic add operation. The non-deterministic order of thread scheduling makes the order of summation non-deterministic.
    
    \item \textbf{Dynamic code paths.} Some deep learning libraries such as cuDNN choose the implementation of an operation according to runtime performance benchmarking, the results of which can be inconsistent from run to run. Different implementations of the operation can use different algorithms, parallelism configurations, data partition configurations, etc., making the computation order non-deterministic.

    \item \textbf{Dynamic batching and caching.} Deep learning inference systems can batch and cache requests dynamically according to the load. As a result, the same request can be included in batches of different sizes from run to run. Then, deep learning libraries may dispatch them to different implementations according to the batch size, and the different implementations result in different computation orders.
\end{itemize}

Non-reproducible factors include non-deterministic factors and the following:

\begin{itemize}
    \item \textbf{Software variability.} Different software or the same software on divergent hardware usually uses different algorithms and computation orders for the same deep learning operation. For example, the computation order of matrix multiplication is diverse and hardware-specific for most libraries such as cuBLAS and Intel MKL.
    
    \item \textbf{Compiler.} Compilers can reorder operations, transform mathematical expressions, and use different arithmetic instructions on different systems. For example, compilers can transform $(a+b)\cdot c$ to $a\cdot c+b\cdot c$, and use fused multiply-add (FMA) instruction instead of multiplication and addition to perform $x\cdot y+z$.
\end{itemize}

\section{RepDL Library}

To address the issue of non-determinism and non-reproducibility in deep learning, we build a library called RepDL, which provides APIs inherited from PyTorch with numerical determinism and reproducibility.

\subsection{Design Principles}

Recall the two types of implementation differences causing the numerical discrepancy: the precision of basic operations and the computation order. RepDL follows two design principles to avoid numerical discrepancies.

\begin{enumerate}
    \item \textbf{Correct rounding for basic operations.} RepDL complies with the correct rounding principle for basic operations as recommended by the IEEE-754 standard \cite{ieee_ieee_2019}. Correct rounding means using the standard IEEE-754 rounding rule to round the infinite-precision real-number result. This principle eliminates the ambiguity in numerical precision.

    \item \textbf{Order invariance for other operations.} RepDL must maintain  order invariance for the operations which are essentially combinations of basic operations. This requires RepDL to implement each operation using the same kind of basic operations with the same order. If an operation has multiple common computation orders, RepDL assigns different APIs to them.
\end{enumerate}

\subsection{Implementation}

\subsubsection{Ensuring correct rounding}

We implement correctly rounded operations to make basic mathematic operations such as arithmetic operations, square root, exponential function and log function reproducible. Common libraries cannot ensure correct rounding according to their documentation. Their hardware-dependent implementations have various precision on different systems. Thus, we develop our correctly rounded operations in RepDL by using correctly rounded mathematical libraries \cite{fousse_mpfr_2007,lim_high_2021} or algorithms with higher precision. As a result, basic operations are ensured to be reproducible.

\subsubsection{Fixing the order of summation}

Unlike basic operations, floating-point summation has no standard definition of correctness. Since floating-point additions are not associative, the result of floating-point summation depends on the order of additions, making reduction operations in deep learning non-deterministic and non-reproducible.

Although floating-point summation can be associative by order-irrelevant summation algorithms \cite{demmel_fast_2013,collange_numerical_2015,ahrens_algorithms_2020}, they are too inefficient for deep learning tasks. In contrast, fixing the order of summations is a more efficient way to achieve reproducibility in practice. We implement two summation orders: (1) sequential summation, which is the default version, and (2) pairwise summation, which is the alternative version and has a different API name. Sequential summation is the most straightforward and cache-friendly implementation, but it may have poor performance when there are insufficient parallel tasks and only a few processors are in use. Pairwise summation can increase parallelism, making it a good alternative for sequential summation.

In deep learning, the fully connected layer and the two-dimensional convolution layer are two types of layers with the intensive floating-point computation. We examine the forward functions of these two types of layers, and find that sequential summation is computationally efficient for them. The analysis for the backward functions is similar.

For a fully connected layer with $N$ input features, $M$ output features, and $B$ batches, there are $t_\mathrm{fc}=B\times M$ independent summation tasks in the forward function, and each task is the summation of $n_\mathrm{fc}=N$ elements. As long as $t_\mathrm{fc}=B\times M$ is greater than the number of processor cores, sequential summation is efficient.

For a convolution layer with $B$ batches, $I$ input channels, $O$ output channels, whose filter shape is $K_w\times K_h$ and output shape is $W\times H$, there are $t_\mathrm{conv}=B\times O \times W\times H$ independent summation tasks in the forward function, and each task is the summation of $n_\mathrm{conv}=I\times K_w \times K_h$ elements. As long as $t_\mathrm{conv}=B\times O \times W\times H$ is greater than the number of processor cores, sequential summation is efficient.

Take ResNet-50 as an example. In ResNet-50, the convolution layers dominate the computational complexity. There are multiple convolution layers whose $t_\mathrm{conv}=B\times 256 \times 56 \times 56 = B\times 802816$. Consider an NVIDIA A100 GPU with 6912 CUDA cores, where the number of cores is far less than $t_\mathrm{conv}$ even if $B=1$. Thus, using sequential summation for convolution layers is efficient.

\subsubsection{Defining the computation graph}

In deep learning libraries like PyTorch, functions are defined by mathematical formulas and can be implemented with various computation orders. This kind of definition implies potential reassociations and transformations that are equivalent in real-number arithmetic. For example, the batch normalization function in PyTorch is defined as $$\mathrm{BatchNorm}(x,\mu,\sigma,w,b,\epsilon) = \frac{x-\mu}{\sqrt{\sigma ^2 +\epsilon}}\cdot w+b.$$
However, for floating-point numbers, the reassociations and transformations can change the result. In the above example, although the documentation of PyTorch defines the batch normalization function as the first line of the formula, it can be implemented with the computation order $$\frac{w}{\sqrt{\sigma ^2 +\epsilon}}\cdot (x-\mu) + b$$ or $$\frac{w}{\sqrt{\sigma ^2 +\epsilon}}\cdot x + \left( b-\frac{\mu}{\sqrt{\sigma ^2 +\epsilon}}\right)$$ by different backend libraries. More precisely, we define the computation order with the computational graph, a directed acyclic graph consisting of basic operations. Therefore, in order to ensure reproducibility, we assign different API names to different computation graph implementations of the same function. 

\subsubsection{Compilation options}

Compilers have optimization options that can reassociate, transform, or reorder mathematical expressions. We disable all the options that cause unsafe mathematical optimizations, except that we enable the floating-point expression contraction option which replaces multiplication and addition with the fused multiply-add (FMA) operation. Modern processors support FMA instructions, with higher precision and performance.

\section{Limitation and Future Work}

Currently, RepDL is not fully optimized. Switching from non-deterministic and non-reproducible deep learning libraries to RepDL can degrade performance mildly. We expect that further performance optimization can mitigate the performance degradation.

RepDL supports the single precision (float32), the default floating-point data type in deep learning. As low-precision floating-point data types become increasingly popular in deep learning, supporting them is the next goal. However, this is more challenging because the arithmetic of low-precision computation is non-standard and hardware-specific, especially on modern GPUs' matrix multiplication units (such as Tensor Cores). We hope that the numerical behavior of low-precision computation will be standardized in the future to facilitate numerical reproducibility.

\section{Conclusion}

This report investigates the root causes of non-determinism and non-reproducibility in deep learning, and presents RepDL, a library designed to enable deterministic and reproducible training and inference. By addressing floating-point computation issues, RepDL provides a foundation for reliable model development and consistent model deployment. 

\bibliographystyle{ACM-Reference-Format}
\bibliography{references}

\end{document}